\documentclass[letterpaper, 10pt, conference]{ieeeconf}  %

\IEEEoverridecommandlockouts                              %
\usepackage{amsmath,amsfonts}
\usepackage{algorithmic}
\usepackage{algorithm}
\usepackage{array}
\usepackage{bbding}
\usepackage{color}
\usepackage{booktabs}
\usepackage{threeparttable}
\usepackage[caption=false,font=normalsize,labelfont=sf,textfont=sf]{subfig}
\usepackage{textcomp}
\usepackage{stfloats}
\usepackage{url}
\usepackage{verbatim}
\usepackage{graphicx}
\usepackage{cite}

\makeatletter
\let\NAT@parse\undefined
\makeatother
\usepackage{hyperref}  

\title{\LARGE \bf DiffAero: A GPU-Accelerated Differentiable Simulation\\ Framework for Efficient Quadrotor Policy Learning}

\author{
Xinhong Zhang, 
Runqing Wang, 
Yunfan Ren, 
Jian Sun, 
\\
Hao Fang, 
Jie Chen,~\IEEEmembership{Fellow,~IEEE},
and Gang Wang$^*$,~\IEEEmembership{Senior Member, IEEE} 
\thanks{This work was supported in part by the National Natural Science Foundation of China under Grants U23B2059, 62173034, 62088101, and also by the Zhongguancun Academy under Grant 20240307.}
\thanks{Xinhong Zhang, Runqing Wang, Jian Sun, Hao Fang, and Gang Wang are with the State Key Lab of Autonomous Intelligent Unmanned Systems, Beijing Institute of Technology, Beijing 100081, China {\tt\small \{xhzhang, bitwrq, sunjian, fangh, gangwang\}@bit.edu.cn}. Xinhong Zhang is also with the Zhongguancun Academy, Beijing 100094, China. Yunfan Ren is with the Department of Mechanical Engineering, University of Hong Kong {\tt\small renyf@connect.hku.hk}. Jie Chen is with the Harbin Institute of Technology, and also with the State Key Lab of Autonomous Intelligent Unmanned Systems, Beijing Institute of Technology, Beijing 100081, China {\tt\small chenjie@bit.edu.cn}.}
\thanks{$^*$Corresponding author.}
}

\begin{document}

\maketitle
\thispagestyle{empty}
\pagestyle{empty}
\allowdisplaybreaks

\begin{abstract}

This letter introduces DiffAero, a lightweight, GPU-accelerated, and fully differentiable simulation framework designed for efficient quadrotor control policy learning. DiffAero supports both environment-level and agent-level parallelism and integrates multiple dynamics models, customizable sensor stacks (IMU, depth camera, and LiDAR), and diverse flight tasks within a unified, GPU-native training interface. By fully parallelizing both physics and rendering on the GPU, DiffAero eliminates CPU-GPU data transfer bottlenecks and delivers orders-of-magnitude improvements in simulation throughput. In contrast to existing simulators, DiffAero not only provides high-performance simulation but also serves as a research platform for exploring differentiable and hybrid learning algorithms. Extensive benchmarks and real-world flight experiments demonstrate that DiffAero and hybrid learning algorithms combined can learn robust flight policies in hours on consumer-grade hardware. The code is available at \tt\small{https://github.com/flyingbitac/diffaero}.

\end{abstract}

\section{Introduction}

Quadrotors—and swarms of quadrotors thereof—are increasingly deployed in complex environments for aerial inspection, environmental monitoring, and high-speed racing, owing to their agile maneuverability and onboard sensing capabilities. Traditional autonomy architectures decompose flight functionality into perception, localization, mapping, planning, and control modules \cite{diffflatvijay, ren2025sciro}. Although modular, this hierarchical pipeline incurs latency and accumulative errors, limiting the full exploitation of quadrotor dynamics \cite{loquercio2021learning}. In addition, the tight coupling between modules renders design and tuning labor-intensive and brittle.

End-to-end learning addresses these limitations by training neural flight policies that map raw sensor observations directly to control commands, thereby streamlining the autonomy stack and enabling tighter feedback loops \cite{kondo2024cgd}.
Reinforcement learning (RL) methods train flight policies through interactions by maximizing expected cumulative reward \cite{zhang2023storm}. However, RL methods, model-free RL in particular, often demand millions of simulated interactions to achieve proficient behavior, especially when processing high-dimensional observations (e.g., depth images) or optimizing sparse rewards \cite{ferede2024end,kaufmann2023champion}.
Imitation learning (IL), on the other hand, reduces sample complexity by leveraging expert demonstrations but suffers from limited generalization and the practical burden of data collection because it depends heavily on demonstration diversity and quality \cite{wang2021robust}.

\begin{figure}[!t]
\centering
\includegraphics[width=0.48\textwidth]{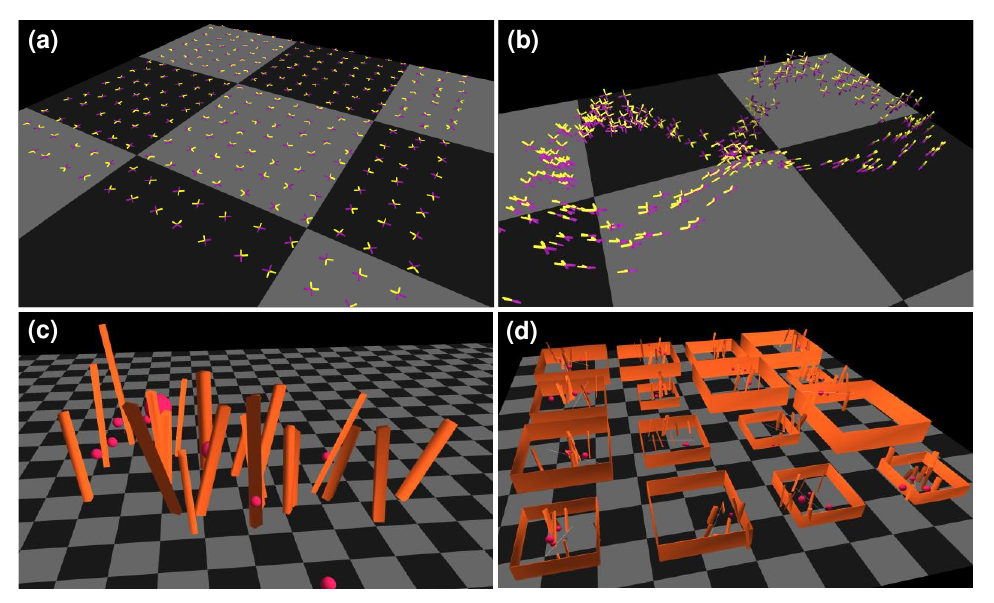}
\vspace{-0.8cm}
\caption{\textbf{Visualization of task environments in DiffAero.} (a) Position control. (b) Racing. (c) Obstacle avoidance with outdoor obstacle settings. (d) Obstacle avoidance task with indoor obstacle settings. Ceilings are omitted for clarity.}\label{head}
\vspace{-0.5cm}
\end{figure}

Accelerating flight policy learning can be approached from two directions: that is improving the simulation performance or enhancing the data efficiency of the learning algorithm. Existing GPU-based drone simulators are typically built upon existing physics and rendering engines, thus limiting their simulation and rendering performance. Furthermore, the lack of differentiability and a unified learning interface hinders the development and benchmarking of novel data-efficient hybrid algorithms across paradigms \cite{xu2024omnidrones, li2024visflyefficientversatilesimulator, huang2025general}.

In this letter, focusing on the autonomous flight of quadrotors, we intend to accelerate quadrotor policy learning by enhancing both simulation performance and data efficiency. We present a GPU-accelerated, \textbf{Diff}erentiable, and extendable simulation framework for \textbf{Aer}ial r\textbf{o}botics, \textbf{DiffAero}, enabling high-performance interactive quadrotor simulation. It offers $1.8\times$ and $9.6\times$ speedups in physics simulation and depth rendering compared to Aerial Gym \cite{kulkarni2025aerial}, respectively. To our best knowledge, this is the first differentiable simulation framework that performs both physical simulation and visual rendering fully on the GPU, as well as supports multiple dynamics models. By combining high-performance simulation, a unified learning interface, learning algorithms, and deployment utilities, robust generalizable flight policies can be trained, evaluated, and deployed within hours on consumer-grade hardware.

In a nutshell, the main contributions of this work are summarized as follows.
\begin{enumerate}
\item We present DiffAero, a GPU-accelerated differentiable quadrotor simulator that parallelizes both physics and rendering. It achieves orders-of-magnitude performance improvements over existing platforms with little VRAM consumption. 
\item DiffAero provides a modular and extensible framework supporting four differentiable dynamics models, three sensor modalities, and three flight tasks. Its PyTorch-based interface unifies four learning formulations and three learning paradigms. This flexibility enables DiffAero to serve as a benchmark for learning algorithms and allows researchers to investigate a wide range of problems, from differentiable policy learning to multi-agent coordination. 
\item We demonstrate how DiffAero serves as an enabling research platform for exploring and evaluating quadrotor learning algorithms. Through simulation and real-world experiments, we show that the framework facilitates the development of data-efficient differentiable and hybrid learning algorithms and seamless sim-to-real transfer, thereby accelerating research on learning-based quadrotor flight control. Moreover, the source code is released for the reference of the community. 
\end{enumerate}

\section{Related Work}

\begin{table*}[!t]
    \caption{Feature Comparison of Popular Drone Simulation Frameworks for Interactive Policy Learning}\label{simulator_comparison}
    \centering
    \resizebox{1.0\textwidth}{!}{
    \begin{tabular}{l|ll|c|cc|c|c|c|ll}
    
        \toprule
    
        & \multicolumn{2}{c|}{\textbf{Engine}}&  \textbf{GPU}& \multicolumn{2}{c|}{\textbf{Supports multiple}}& \textbf{Sensor}& \textbf{Is}&\textbf{Deploy}& \multicolumn{2}{c}{\textbf{FPS}}\\
        & \textbf{physics}& \textbf{rendering}&  \textbf{parallelization}&  \textbf{robot configurations}&\textbf{dynamics models}& \textbf{modalities}$^1$& \textbf{differentiable}& \textbf{utilities}& \textbf{physics}$^2$&\textbf{depth rendering}$^3$\\
    
        \midrule
        
        Flightmare$^4$& Flexible& Unity&  \textcolor{red}{\XSolidBrush}&  \textcolor{red}{\XSolidBrush}&\textcolor{red}{\XSolidBrush}& RGB, D, I& \textcolor{red}{\XSolidBrush}&\textcolor{red}{\XSolidBrush} & $2.2\times10^5$&$2.3\times10^2$\\
        
        OmniDrones& PhysX& Omniverse RTX&  \textcolor{green}{\Checkmark}&  \textcolor{green}{\Checkmark}&\textcolor{red}{\XSolidBrush}& RGB, D, I, L& \textcolor{red}{\XSolidBrush}&\textcolor{red}{\XSolidBrush} & $8.4\times10^5$&-\\
        VisFly& Self-defined& Habitat-Sim&  (\textcolor{green}{\Checkmark})&  \textcolor{red}{\XSolidBrush}&\textcolor{red}{\XSolidBrush}& RGB, D, I& \textcolor{green}{\Checkmark}&\textcolor{red}{\XSolidBrush} & $9\times10^4$&$7\times10^3$\\
        Aerial Gym& PhysX& Warp&  \textcolor{green}{\Checkmark}&  \textcolor{green}{\Checkmark}&\textcolor{red}{\XSolidBrush}& RGB, D, I, L& \textcolor{red}{\XSolidBrush}&\textcolor{red}{\XSolidBrush} & $1.9\times10^6$&$1\times10^4$\\
        
        \midrule
        
        \textbf{DiffAero} (Ours)& Self-defined& Self-defined&  \textcolor{green}{\Checkmark}&  \textcolor{green}{\Checkmark}&\textcolor{green}{\Checkmark}& D, I, L& \textcolor{green}{\Checkmark}&\textcolor{green}{\Checkmark} & $\mathbf{3.4\times10^6}$&$\mathbf{9.6\times10^4}$\\
        
        \bottomrule
        
    \end{tabular}
    }
    \begin{tablenotes}
    \footnotesize
    \item[]$^1$D stands for depth camera, I for IMU, and L for LiDAR. $^2$FPS results of physics simulation are measured under $8,192$ parallel environments. $^3$FPS results of depth camera rendering are measured under $2,048$ parallel environments with a depth image resolution of $64\times64$, except OmniDrones, which does not support depth cameras. $^4$Data from \cite{li2024visflyefficientversatilesimulator}. Note that the GPU parallelization support of VisFly is indicated with a (\textcolor{green}{\Checkmark}), as rendered images in VisFly are returned as NumPy arrays stored in RAM. Consequently, its simulation performance is significantly limited by the I/O speed.
    \end{tablenotes}
\vspace{-0.5cm}
\end{table*}

\subsection{Learning-based Autonomous Flight}
Learning-based end-to-end policies enable direct control from raw observations, bypassing the limitations of traditional modular pipelines. Imitation learning approaches supervise policy networks to mimic expert demonstrations across a variety of tasks, including acrobatics \cite{kaufmann2020RSS}, navigation \cite{loquercio2021learning} and racing \cite{wang2021robust}, but their performance relies heavily on diverse demonstrations and lacks generalization \cite{Zhang2025}. RL allows training flight policies interactively using simulation generated data and rewards, demonstrating champion-level results in drone racing \cite{kaufmann2023champion}.

While most RL methods treat quadrotor dynamics as a black box, recent advances integrate differentiable simulation to incorporate dynamics as analytic gradients into the learning process, enabling more data-efficient training and direct optimization over visual features. Differentiable simulation approaches \cite{Zhang2025} enabled end-to-end agile navigation of a swarm of quadrotors from depth images. While these approaches are data-efficient and capable of handling high-dimensional inputs, designing effective differentiable reward signals remains challenging. Furthermore, similar to artificial potential field methods, learning agents can easily become trapped in local optima. Hybrid algorithms like SHAC \cite{xu2021accelerated} enable the integration of both RL and differentiable learning techniques, allowing for the separate and flexible design of reward signals for direct gradient backpropagation and terminal value optimization.

\subsection{Drone Simulators}
Early drone simulators like Gazebo \cite{koenig2004design} became popular due to ROS integration and flexible support for physics and sensor simulation, but they face limitations in performance and scalability from a data generation perspective. With a primary emphasis on visual realism, Flightmare \cite{song2021flightmare} and AirSim \cite{shah2018airsim} leveraged game engines to provide fast simulation with realistic environment rendering, while Menagerie \cite{menagerie2022github} and PyBulletDrones \cite{panerati2021learning} targeted high-fidelity physics based on Mujoco \cite{todorov2012mujoco} and PyBullet \cite{coumans2016pybullet}, though their visual realism remains of low quality.

Recent simulators, such as OmniDrones \cite{xu2024omnidrones}, Aerial Gym Simulator \cite{kulkarni2025aerial} and AirGym \cite{huang2025general} utilize NVIDIA Omniverse Isaac Sim and Isaac Gym \cite{makoviychuk2021isaac} to explore GPU-acceleration for parallel simulation and rendering, minimizing CPU-GPU data transfer by directly accessing simulator states as PyTorch \cite{paszke2019pytorch} tensors. However, these frameworks are resource-intensive, demanding strict system requirements, and the rasterization pipeline still limits their rendering performance. VisFly \cite{li2024visflyefficientversatilesimulator} incorporates differentiable simulation, but its performance is bottlenecked by the I/O speed due to the image transfer between RAM and VRAM, and does not provide a unified interface that is compatible with both RL and differentiable learning algorithms. To overcome these issues, we develop a novel simulation framework from scratch entirely in PyTorch, providing fully GPU-parallelized differentiable simulation capabilities while remaining lightweight, flexible, easy to install, and user-friendly. 

\section{The DiffAero Simulation Framework}

\begin{figure*}[!ht]
\centering
\includegraphics[width=1.01\textwidth]{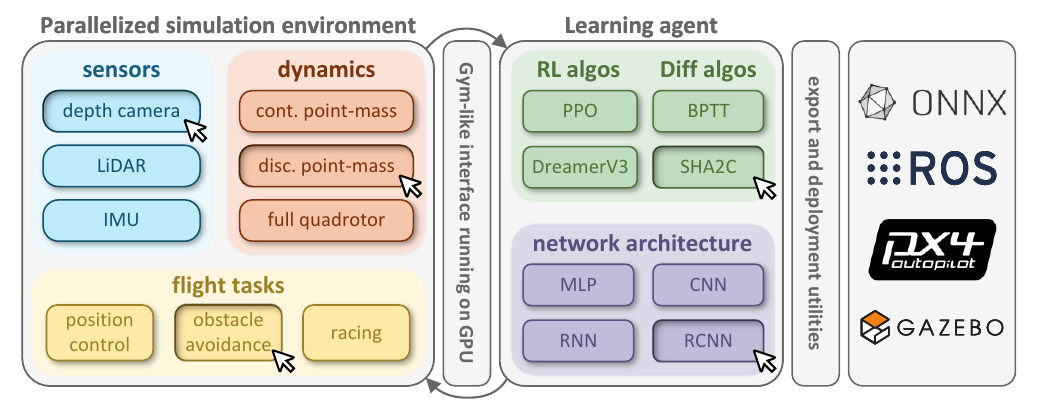}
\vspace{-0.8cm}
\caption{\textbf{An overview of the developed simulation framework.} The simulation framework comprises three modular components: (1) simulation environments, (2) learning agents, and, (3) export and deployment utilities. The decoupled modular design of simulation environment and learning agent enables nearly arbitrary module combinations, thereby providing highly customizable training configurations. Additionally, DiffAero features a Gym-like learning interface that runs on the GPU, allowing data to be transferred directly between environments and agents, thereby reducing data transfer I/O overhead. Furthermore, DiffAero incorporates export and deployment utilities to deploy and evaluate the trained policy.}\label{framework}
\vspace{-0.5cm}
\end{figure*}

In this section, we describe the key features and building blocks of DiffAero. DiffAero is a differentiable, high-performance quadrotor simulation toolkit capable of simulating multiple environments in parallel, each containing multiple quadrotor instances, by leveraging the parallel computing capabilities of modern GPUs.

DiffAero consists of the following main components: (1) a high-performance and flexible simulation framework, including three quadrotor dynamics, three sensor modalities, and three flight tasks, combined with a Gym-like interactive learning interface; (2) an interactive learning library implemented in PyTorch \cite{paszke2019pytorch}, including four configurable network architectures and several learning algorithms, optimized for training on data generated by GPU-parallelized simulation; and, (3) utilities to export and deploy trained control policies in other simulators or on real quadrotor platform, enable researchers to evaluate the performance efficiently. An overview of DiffAero is depicted in Fig. \ref{framework}.

\subsection{Differentiable Dynamics}
Existing differentiable physics engines \cite{brax2021github, warp2022} primarily support general rigid-body dynamics and use joint torques as action inputs, which mismatches the requirements of quadrotor simulation. Instead of pursuing generality, our framework is tailored specifically for quadrotors by implementing four dynamics models with varying levels of complexity and fidelity to facilitate parallel, GPU-accelerated, and multi-fidelity quadrotor simulation.

The first dynamics model is the full quadrotor dynamics \cite{zhou2023efficient}, defined as follows
\begin{equation}\label{fullquad}
\begin{cases}\begin{aligned}
\dot{\mathbf{p}}=&\mathbf v \\
\dot{\mathbf{v}}=&\mathbf g-c\mathbf z_B-\mathbf {RDR^\top v} \\
\dot{\mathbf{q}}=&\frac{1}{2}\mathbf q\odot\begin{bmatrix}0\\\mathbf\omega\end{bmatrix} \\
\dot{\mathbf{\omega}}=&\mathbf J^{-1}(\mathbf{\tau-\omega\times J\omega})
\end{aligned}\end{cases}
\end{equation}
where $\mathbf p$ and $\mathbf v$ are the position and velocity in world frame, $\mathbf q$ and $\mathbf\omega$ denote the orientation quaternion and angular velocity in body frame, $\mathbf g$ indicates the gravitational acceleration vector, $\mathbf z_B$ refers to the unit vector along the z-axis of the body frame, $\mathbf R$, $\mathbf D$ and $\mathbf J$ are respectively the rotation matrix, drag matrix, and inertia matrix, defined in body frame, $c$ and $\tau$ represent collective rotor thrust and torque, and $\odot$ signifies the quaternion product. We use the geometric controller \cite{lee2010control} that takes thrust and body rate commands as inputs to generate desired collective thrust and torque. The dynamics model \eqref{fullquad} provides the highest fidelity but is also the most computationally demanding; its complex computation graph makes it suitable primarily for reinforcement learning tasks where precise attitude dynamics are critical.

The simplified quadrotor dynamics \cite{heeg2025learning} is defined as follows
\begin{equation}\label{simplequad}
\begin{cases}\begin{aligned}
\dot{\mathbf{p}}=&\mathbf v \\
\dot{\mathbf{v}}=&\mathbf Rc+\mathbf g\\
\dot{\mathbf{R}}=&\mathbf R[\mathbf \omega]_\times
\end{aligned}\end{cases}
\end{equation}
where $[\cdot]_\times$ denotes the skew-symmetric matrix operator. The dynamics model \ref{simplequad} bypasses body-rate dynamics by directly taking body rates and collective thrust as inputs while preserving attitude dynamics.

The remaining two dynamics models are based on the point-mass dynamics for differentiable learning algorithms. The continuous-time point-mass dynamics is defined as follows
\begin{equation}\label{pmc}
\begin{cases}
\dot{\mathbf{p}}=\mathbf v \\
\dot{\mathbf{v}}=\mathbf a+\mathbf g - d\mathbf v \\
\dot{\mathbf{a}}=\lambda(\mathbf u - \mathbf a)
\end{cases}
\end{equation}
where $\mathbf{a}$ represents the acceleration of the quadrotor in the world frame, $d$ is the drag coefficient, and $\lambda$ is the control latency factor. The dynamics model \eqref{pmc} additionally omits the rotational degrees of freedom compares to \ref{simplequad} and primarily focuses on the movement of the center of mass, thereby simplifying the computation and enabling smooth gradient backpropagation through the dynamics. Following \cite{Zhang2025}, the discrete-time point-mass dynamics is defined as follows
\begin{equation}\label{pmd}
\begin{cases}
\mathbf p_{t+1}=\mathbf p_t + \mathbf v_t\Delta t+\dfrac{1}{2}\mathbf u_t\Delta t^2 \\
\mathbf v_{t+1}=\mathbf v_t + \dfrac{1}{2}(\mathbf u_t + \mathbf u_{t+1})\Delta t
\end{cases}
\end{equation}
The dynamics model \eqref{pmd} is the simplest among all implemented models and requires minimal computation, making it suitable for tasks prioritizing extreme simulation speed.

The dynamics models described above are implemented in PyTorch, thereby supporting direct gradient backpropagation, single- and multi-agent setups, and massively parallel simulations on GPUs. Given a state $\mathbf s_{t_2}$ and an action $\mathbf a_{t_1}$ with $t_2 > t_1$, the gradient of $\mathbf s_{t_2}$ with respect to $\mathbf a_{t_1}$ can be computed as follows through back-propagation
\begin{equation}\label{backprop}
\frac{\partial \mathbf s_{t_2}}{\partial \mathbf a_{t_1}}=\frac{\partial \mathbf s_{t_1+1}}{\partial \mathbf a_{t_1}}\prod^{t_2-1}_{t=t_1+1}\frac{\partial \mathbf s_{t+1}}{\partial \mathbf s_t}
\end{equation}
Although primarily designed for quadrotors, these models can be readily adapted to other multi-rotor configurations (e.g., hexarotors and octarotors) with minor modifications. DiffAero enables further research on dynamics modeling by making all dynamics modular, supporting them in all flight tasks, and offering a unified base class for custom dynamics.

\subsection{Sensor Stack}
\begin{figure}[!t]
\centering
\includegraphics[width=0.3\textwidth]{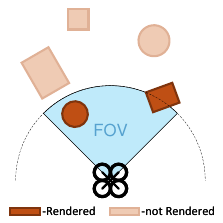}
\vspace{-0.2cm}
\caption{\textbf{Rendering strategy of ray-casting sensors.} Obstacles outside the sensor’s field of view are excluded from the ray-casting process, substantially accelerating rendering.}
\label{fov}
\vspace{-0.5cm}
\end{figure}

To enable end-to-end flight policy learning, DiffAero provides interceptive and exteroceptive sensor data regarding the agent and its surrounding obstacles. DiffAero supports three sensor modalities: IMU, depth camera, and LiDAR, all implemented in PyTorch. Both depth camera and LiDAR are implemented with a self-defined ray-casting algorithm. At a resolution below $64\times64$, which is crucial for agents to learn generalizable policies from simple obstacle configurations and visual effects, GPU-parallelized ray-casting offers significant performance and usability advantages compared to rasterization pipelines. By excluding obstacles outside the sensor’s FOV from the ray casting pipeline, the rendering efficiency is greatly improved, particularly in cluttered environments, as shown in Fig. \ref{fov}. Additionally, rather than performing computationally expensive ray-casting against individual triangle meshes \cite{kulkarni2025aerial,Zhang2025}, we implement specialized ray-casting functions for every primitive shape (e.g., spheres, cubes, cylinders) to further accelerate the rendering pipeline.

DiffAero also provides access to the quadrotor’s IMU data with configurable noise and drift parameters for odometry-based policy learning. Following \cite{hu2024seeing}, the attitude of the point-mass dynamics model \eqref{pmc} and \eqref{pmd} are governed by the collective thrust direction and velocity vector. Specifically, since propeller forces are aligned with the body frame’s z-axis, two rotational degrees of freedom are determined by the collective thrust vector. The remaining degree of freedom is constrained by aligning the body frame’s x-axis with the projection of the exponential moving average of the velocity vector onto the horizontal plane. This configuration ensures that the quadrotor consistently orients itself in the direction of motion, reducing the task difficulty and avoiding sideslip, which can be problematic sometimes because of inadequate observation about the environment. However, it is important to note that this approach might be less effective in highly cluttered or indoor environments, where rotational maneuvers in place are crucial for gathering detailed environmental information.

\begin{figure*}[!ht]
\centering
\includegraphics[width=1.01\textwidth]{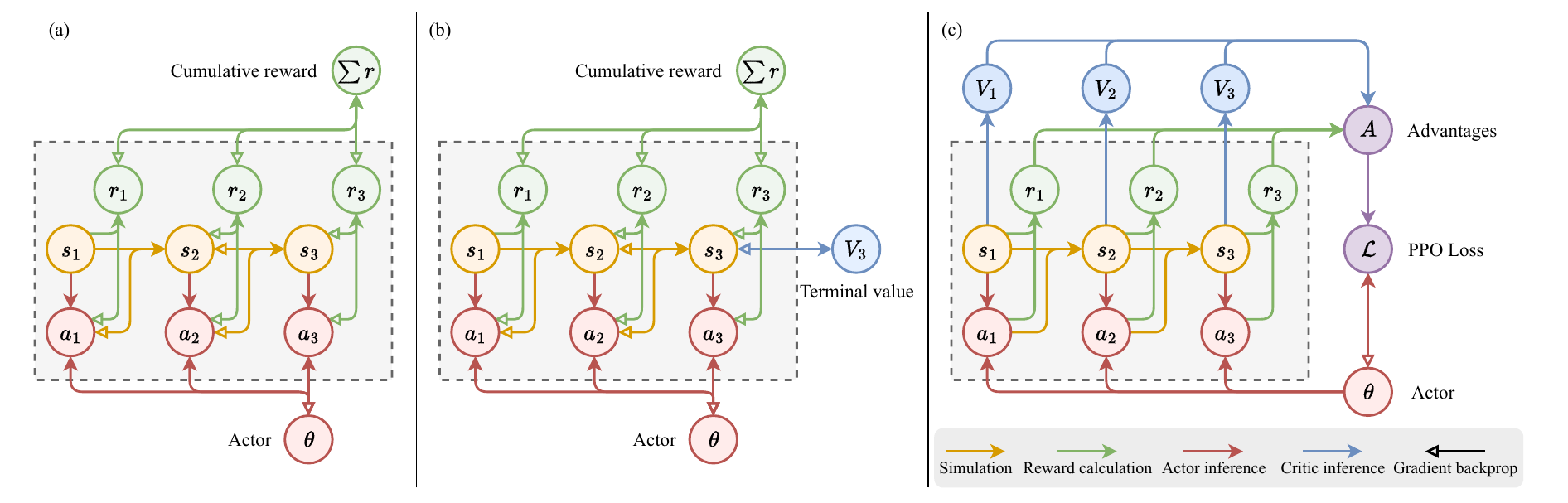}
\vspace{-0.8cm}
\caption{\textbf{Different learning paradigms under a unified learning interface.} (a) Differentiable learning algorithms, (b) Hybrid learning algorithms, and (c) Reinforcement learning algorithms are all supported by the interface.}
\label{paradigms}
\vspace{-0.5cm}
\end{figure*}

\subsection{Flight Tasks}
For high-speed autonomous flight, we implement three flight tasks.
\begin{itemize}
    \item \textbf{Position Control}: The quadrotor(s) must reach and hover at specified target positions from random initial states. This task primarily evaluates the fidelity of the dynamics models and the effectiveness of learning algorithms. Under the multi-agent configuration, quadrotors additionally maintain a prescribed formation while avoiding inter-agent collisions.
    \item \textbf{Obstacle Avoidance}: The quadrotor(s), equipped with a depth camera or a LiDAR, navigate to and hover on target positions while avoiding collision with environmental obstacles. The obstacles are randomly placed around the path between the initial position and the target position, ensuring a non-trivial obstacle avoidance policy. Formation and inter-agent collision-avoidance constraints remain active in the multi-agent setting.
    \item \textbf{Autonomous Racing}: The quadrotor traverses through a set of gates in a predefined order as quickly as possible. Gate positions are provided either through onboard depth sensing or as ground-truth relative poses.
\end{itemize}

\begin{figure}[!b]
\centering
\vspace{-0.5cm}
\includegraphics[width=0.45\textwidth]{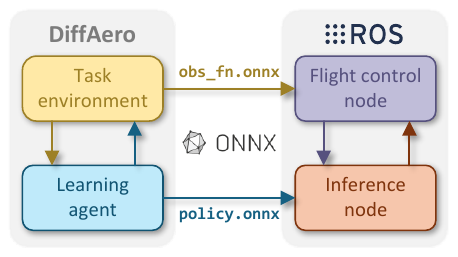}
\vspace{-0.3cm}
\caption{\textbf{Training and deployment pipeline.} The observation function and the learned policy are exported from DiffAero in ONNX format and loaded by the ROS flight control and inference nodes.}
\label{export}
\end{figure}

The task environments are visualized in Fig. \ref{head}. Both position control and obstacle avoidance tasks support single- and multi-agent configurations. Observations consist of proprioceptive states, goal-related vectors, and exteroceptive measurements (depth or LiDAR). A copy of the flight environment of each task is provided in Gazebo. Once a policy is trained, both the policy and the definition of the observation space are exported to ONNX format, loaded by ROS nodes, and used for inference during SITL or real-world experiments, as illustrated in Fig.~\ref{export}. Packaging the inference function and observation specification in ONNX enables seamless evaluation and deployment of trained policies.

\subsection{Learning Interface}
DiffAero provides a unified Gym-like learning interface for RL and differentiable simulation based learning algorithms. By providing both differentiable and non-differentiable reward signals for each time step, DiffAero enables easy development, evaluation, and comparison of RL and differentiable learning algorithms.
Instead of using existing RL algorithm libraries like SB3 \cite{stable-baselines3} and rsl\_rl \cite{rudin2022learning}, we implement common backbone networks for interactive learning from scratch in PyTorch, including MLP, CNN, RNN, and RCNN, since existing libraries are not capable of leveraging system differentiability. Additionally, we implement several learning agents with unified interfaces, decoupled from specific neural network design, allowing easy comparison of different network architectures.

DiffAero supports four learning formulations: single-agent reinforcement learning (SARL), multi-agent reinforcement learning (MARL), single-agent differentiable learning, and multi-agent differentiable learning. Beyond RL and differentiable learning, hybrid learning algorithms that combine training techniques of both are also supported, as depicted in Fig. \ref{paradigms} (b). With the support of the versatile interface, we implement several learning algorithms, including PPO \cite{schulman2017proximal}, MAPPO \cite{yu2022surprising}, BPTT, SHAC \cite{xu2021accelerated}, multi-agent SHAC, and DreamerV3 \cite{hafner2025mastering}, etc. Since network architecture, learning agent, and algorithmic logic are decoupled from each other, users can readily develop custom algorithms without concerning themselves with low-level implementation details.

\subsection{Designing Novel Algorithms with DiffAero}
To illustrate how DiffAero enables the development and evaluation of new algorithms and learning paradigms, we present a simple case study on training a vision-based navigation policy with a novel hybrid learning algorithm. We designed a variant of SHAC \cite{xu2021accelerated} by using an asymmetric actor-critic architecture and different reward signals for direct gradient back-propagation and terminal value optimization, respectively, formulating short horizon asymmetric actor-critic (SHA2C).

Specifically, the policy network $\pi_\theta$ adopts a recurrent and convolution architecture to encode visual inputs and preserve temporal information, while the value network $V_\phi$ digests the full simulator state with an MLP. We design two reward signals for the obstacle avoidance task: a local control reward function $R_{\text{ctrl}}$ and a goal-oriented reward function $R_{\text{goal}}$. Both reward signals are designed to guide the quadrotor to navigate to the target and avoid collision, where the control reward function $R_{\text{ctrl}}$ is dense and differentiable to the physical state of the agent, controlling the short-term behavior of the agent, and the goal-oriented reward function $R_{\text{goal}}$ is sparse and non-differentiable, supervising the long-term and global behavior with binary success and failure flags, and representing the overall goal of the task.

At each time step, the policy network $\pi_\theta$ selects actions $\mathrm a_t\sim\mathcal N(\mu_\theta( \mathrm o_t),\sigma_\theta( \mathrm o_t))$, while the value network $V_\phi$ evaluates the state value of $R_{\text{goal}}$ from the full physical state $\mathrm s_t$. After collecting $N$ rollouts of length $T$, SHA2C updates both networks in an on-policy manner, using data collected by the agent. Specifically, the policy loss is computed by accumulating discounted local rewards and a terminal value as follows
\begin{equation}\label{actor_loss}
    L(\theta)=-\frac{1}{NT}\sum_{i=1}^{N}\left[\Big(\sum^{T-1}_{t=0}\gamma^tR_{\text{ctrl}}(\mathrm s^i_t,\mathrm a^i_t)\Big)+\gamma^TV_\phi(\mathrm s^i_{T})\right]
\end{equation}
where $\mathrm s^i_t$ and $\mathrm a_t^i$ are the state and action at step $t$ of the $i$-th trajectory. The value network is updated using the MSE loss, matching its outputs to the value targets bootstrapped using the $k$-step return of the goal reward $R_{\text{goal}}$ from time $t$ through TD-$\lambda$.

Since SHA2C optimizes $R_\text{ctrl}$ and $R_\text{goal}$ with first- and zeroth-order policy gradient, respectively, it relaxes the constraint that all components of the reward signal in differentiable simulation based learning algorithms must be dense, continuous, and differentiable. With proper design, the policy learning process benefits from $R_\text{goal}$ even when it is simple and sparse. This example shows that the versatility of DiffAero enables researchers to develop and evaluate novel algorithms and learning paradigms.

\section{Experiments}

To evaluate the performance of the simulation framework and demonstrate the effectiveness of the proposed algorithm SHA2C, in this section we present experimental results on: (1) simulation speed and VRAM utilization; (2) performance of baseline algorithms in flight tasks; (3) comparison of different quadrotor dynamics; and, (4) deploy examples of flight policies trained by SHA2C.
\begin{figure}[!t]
\centering
\includegraphics[width=0.5\textwidth]{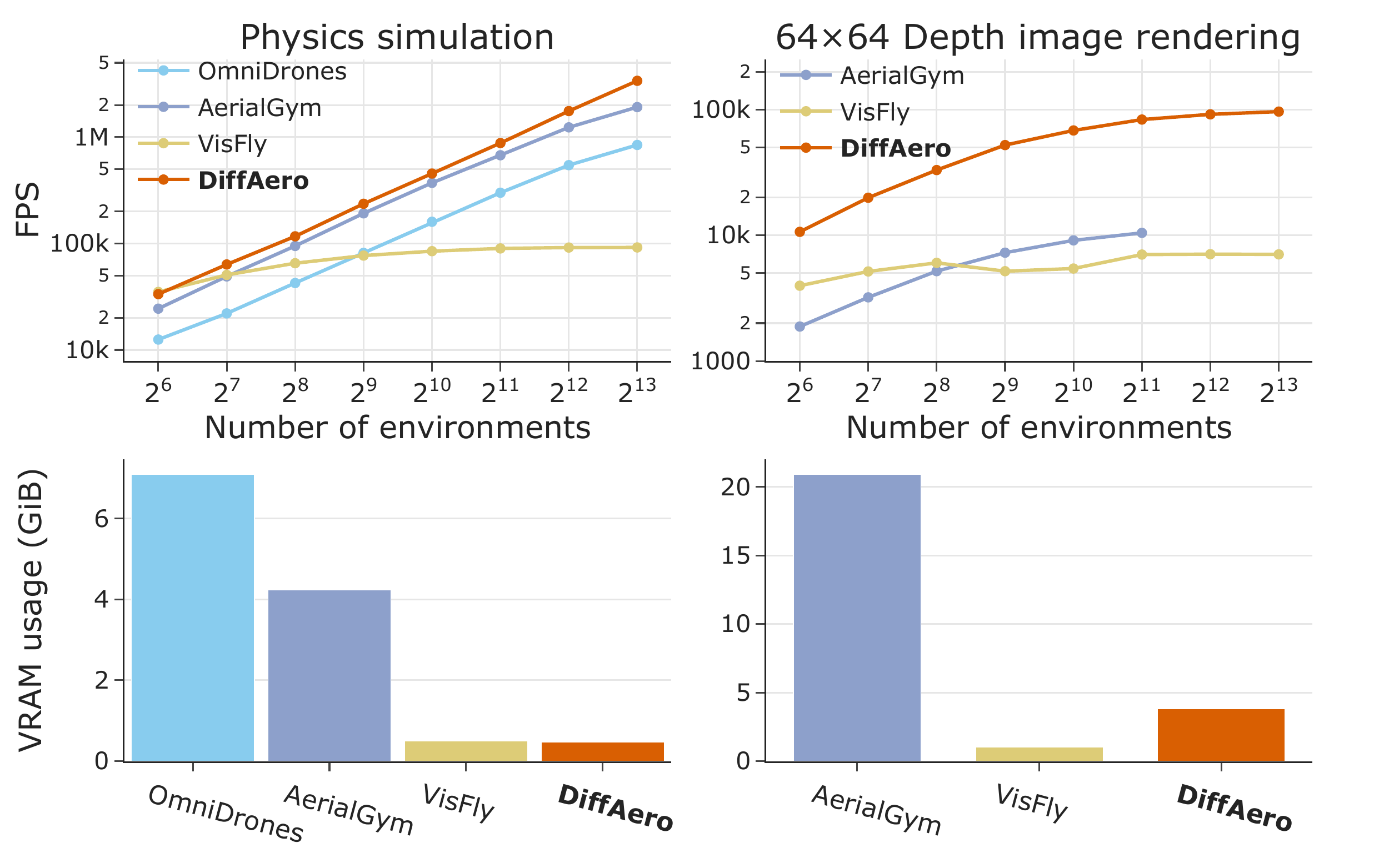}
\vspace{-0.8cm}
\caption{\textbf{Comparison of the simulation performance among GPU-parallelized drone simulators.} The upper two figures depict FPS of simulators when performing physics simulation (total number of interactions per second) and rendering $64\times64$ depth images (total number of images per second) under varying numbers of environments. The lower two figures present the VRAM consumption of simulators during physics simulation and depth rendering, evaluated using a fixed number of $2,048$ parallel environments. Note that the depth rendering performance of OmniDrones is not provided since it does not support depth camera functionality.}
\label{performance}
\vspace{-0.5cm}
\end{figure}

\subsection{Simulation Performance}
We compared the simulation speed and VRAM consumption of several simulators designed exclusively for drones, as illustrated in Fig. \ref{performance}. The results indicate that our simulator achieves superior performance in both physics simulation and depth image rendering. Additionally, it maintains VRAM consumption within a practical range, enabling high-speed simulations on mainstream consumer-grade hardware. Notably, the physics simulation speed of our simulator continues to scale effectively without saturation, even with a massive number of environments (e.g., $8,192$), whereas other simulators exhibit saturation. This demonstrates the scalability potential of our simulation framework. All reported results incorporate the time and VRAM overhead associated with observation calculation, reward calculation, and environment reset operations. All results were obtained on a desktop workstation equipped with an Intel Core i9-14900K processor and an NVIDIA GeForce RTX 4090 graphics card.

\begin{figure}[!b]
\vspace{-0.5cm}
\centering
\includegraphics[width=0.5\textwidth]{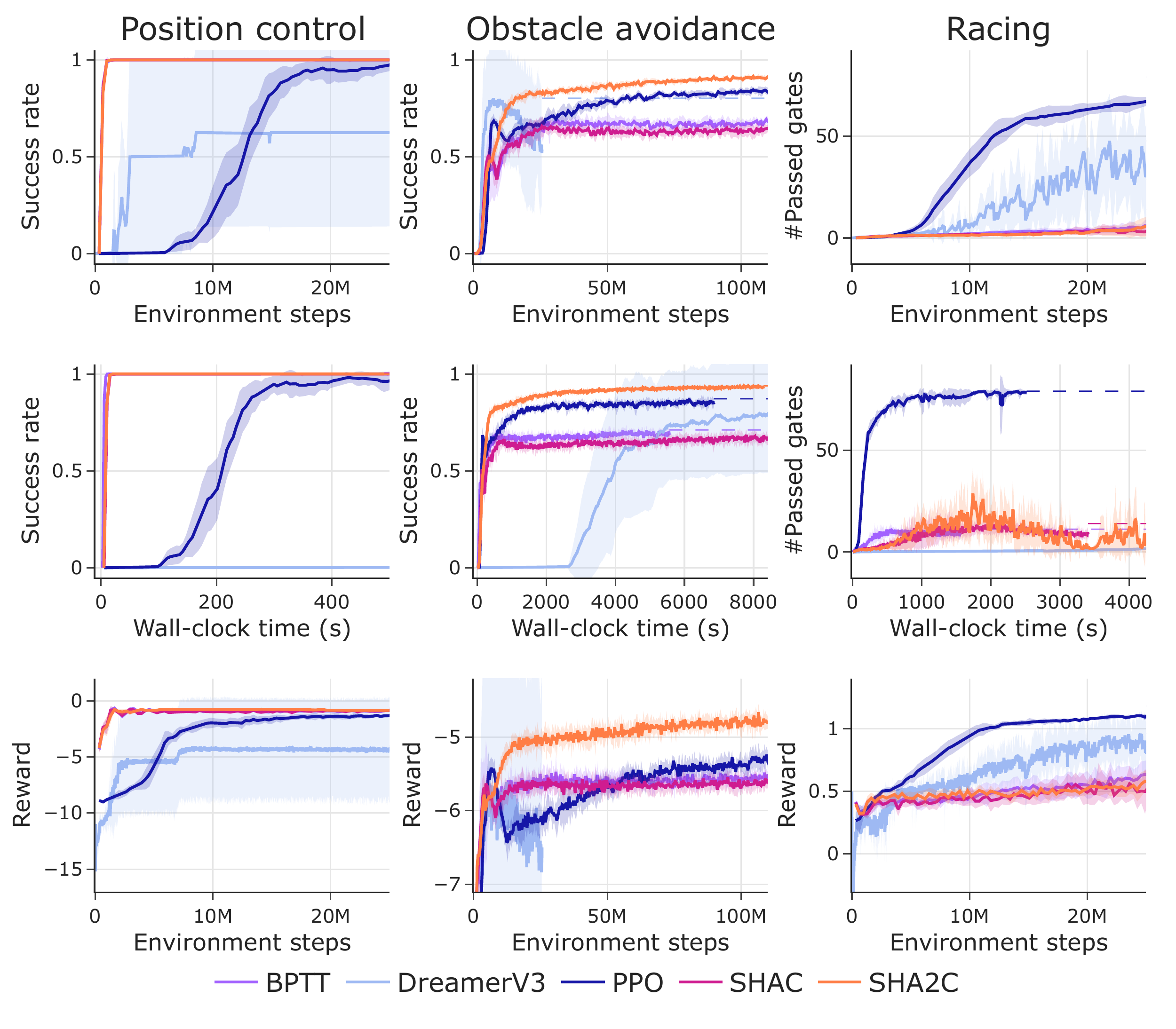}
\vspace{-0.8cm}
\caption{\textbf{The learning curves of algorithm baselines on flight tasks provided by the simulation framework.} All results are obtained using continuous point-mass dynamics, follow the same number of parallel environments and number of policy updates, except DreamerV3, which needs far more updates and fewer data than other algorithms.}
\label{baseline}
\end{figure}

\subsection{Benchmark Flight Tasks}
We tested the performance of algorithm baselines in single-agent versions of flight tasks. The results are shown in Fig. \ref{baseline}. All evaluated algorithms were adapted from open-source implementations and modified to interface seamlessly with our learning agents and network architectures. For each algorithm, the same set of hyperparameters was applied across all tasks, except network architecture, which was tailored for each algorithm to achieve the best performance. The weights assigned to individual reward components for RL algorithms and differentiable algorithms were tuned separately, since RL algorithms are sensitive to the magnitude of the reward signal, whereas differentiable algorithms are affected primarily by the derivative of the reward, regardless of its value. The reward results reported in Fig. \ref{baseline} are weighted in the same manner as the differentiable reward, except for the Racing task, in which we report the reward signal for RL algorithms.

As shown in Fig. \ref{baseline}, in the position control task, differentiable algorithms consistently outperformed RL algorithms, achieving even greater data efficiency than the model-based DreamerV3. We attribute this to the simplicity of the point-mass dynamics model and the goal of the task, which makes the differentiable reward signals both intuitive and effective. All algorithms successfully converged to a 100\% success rate in this task. In the obstacle avoidance task, SHA2C achieved the best success rate and stability among the algorithms we tested. Compared to BPTT and SHAC, though using the same differentiable reward signal $R_\text{ctrl}$, SHA2C was able to further improve the policy under the guidance of $R_\text{goal}$, avoiding convergence to local optima. For the racing task, we adopted the design of the progress reward following \cite{kaufmann2023champion}. Since we've failed to design an effective differentiable reward signal to guide the agent smoothly through the gates, all differentiable learning algorithms exhibited poor performance in this task.

\begin{figure}[!t]
\centering
\includegraphics[width=0.48\textwidth]{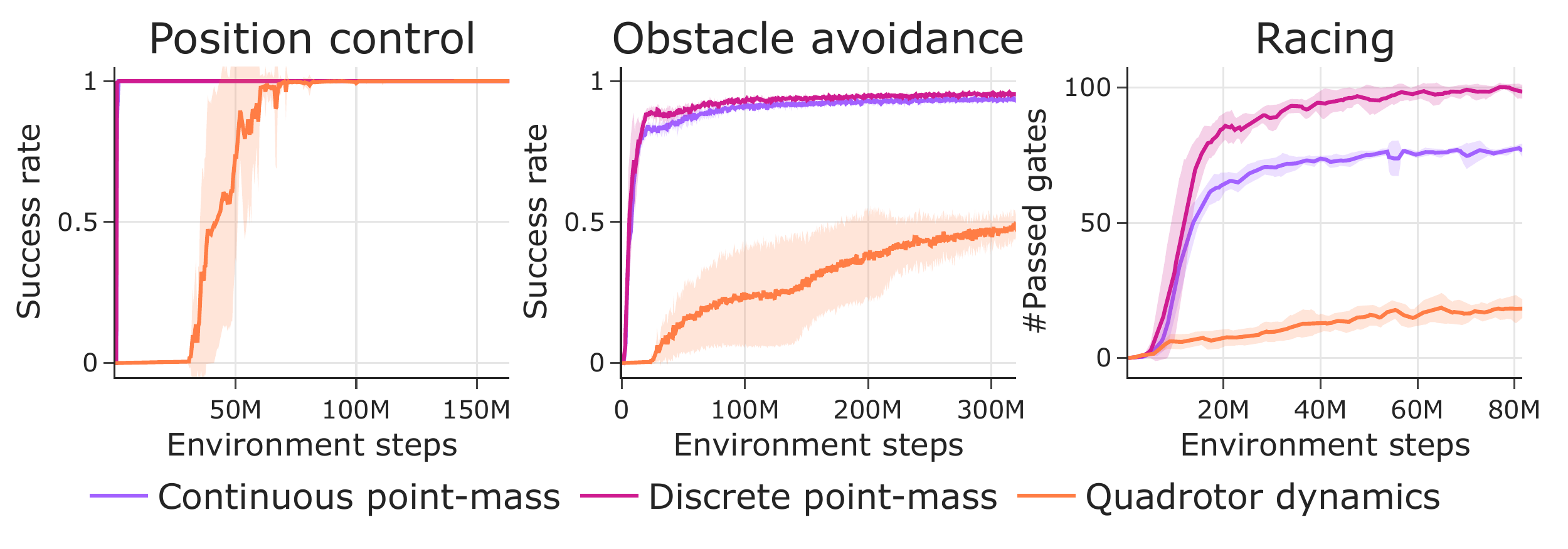}
\vspace{-0.8cm}
\caption{The learning curves of different dynamics models on flight tasks.}
\label{dynamics}
\vspace{-0.5cm}
\end{figure}

\subsection{Comparison of Dynamics Models}
We compared the learning curves of different dynamics models across the three flight tasks provided in our framework, as shown in Fig. \ref{dynamics}. For each task, the policy was trained using the algorithm that achieved the best performance on the corresponding task (i.e., SHA2C for position control and obstacle avoidance, PPO for racing). The results highlight a clear trade-off: full quadrotor dynamics provides higher fidelity by retaining detailed attitude information but incurs substantial training overhead and higher difficulty, leading to slower convergence. In contrast, point-mass dynamics models sacrifice attitude fidelity yet dramatically simplify the optimization landscape, resulting in faster and more stable policy learning.

\begin{figure}[!t]
\centering
\includegraphics[width=0.48\textwidth]{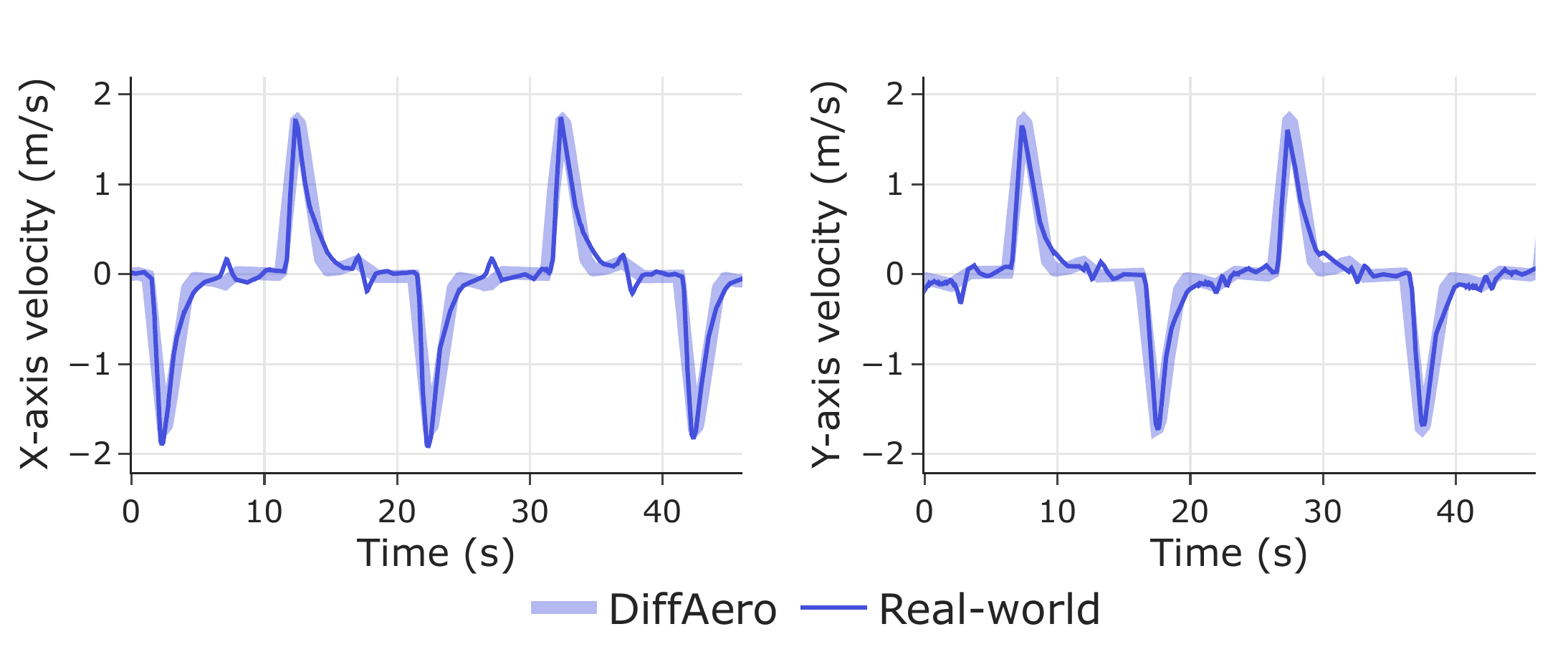}
\vspace{-0.8cm}
\caption{Velocity curve comparison between DiffAero and the real world under the position control scenario.}
\label{deploy_pc}
\vspace{-0.5cm}
\end{figure}

\begin{figure}[!b]
\vspace{-0.5cm}
\centering
\includegraphics[width=0.48\textwidth]{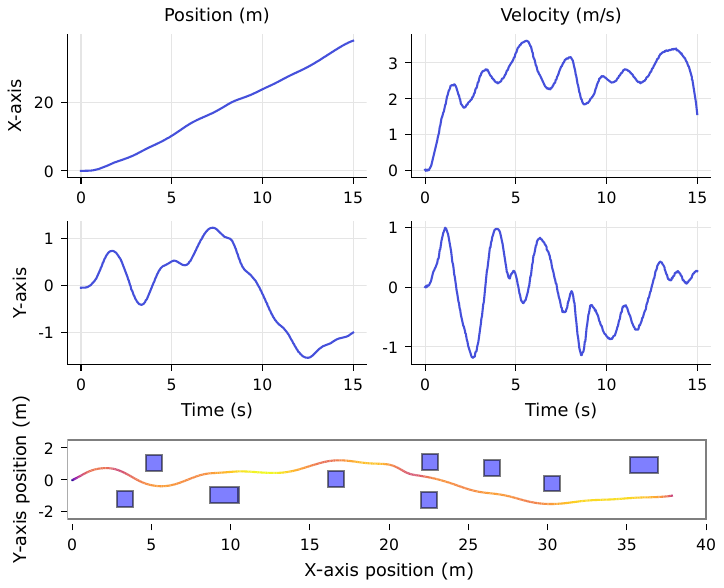}
\vspace{-0.8cm}
\caption{Trajectory of vision-based obstacle avoidance in Gazebo.}
\label{deploy_oa}
\end{figure}

\subsection{Deployment Examples}
In this section, we present the deployment results of our trained policies both in Gazebo and in real-world experiments. The policies were trained using the proposed SHA2C algorithm in $1,024$ environments in parallel, using continuous point-mass dynamics. Desired accelerations generated by the agent were converted into target attitude and thrust commands and fed into the PX4 flight control stack. Air drag coefficient $d$, control latency $\lambda$, and action range were randomized at the start of each episode for generalization, and the randomization range of $\lambda$ was empirically calibrated to match the dynamics of the real quadrotor \cite{RLsim2real}. To learn a yaw-invariant flight policy, observations and actions are defined in local frame that retains only yaw attitude while discarding pitch and roll angles.

For real-world experiments, we employed an OptiTrack motion capture system to estimate the quadrotor’s position, velocity, and attitude. Onboard computation is handled by a Radxa X4 computer equipped with an Intel N100 processor. Depth images are captured by a RealSense D435i camera and post-processed with filters provided by the RealSense SDK, and downsampled to $16\times9$ pixels, suppressing noise and minimizing the sim-to-real gap of visual features.

\begin{figure}[!t]
\centering
\includegraphics[width=0.44\textwidth]{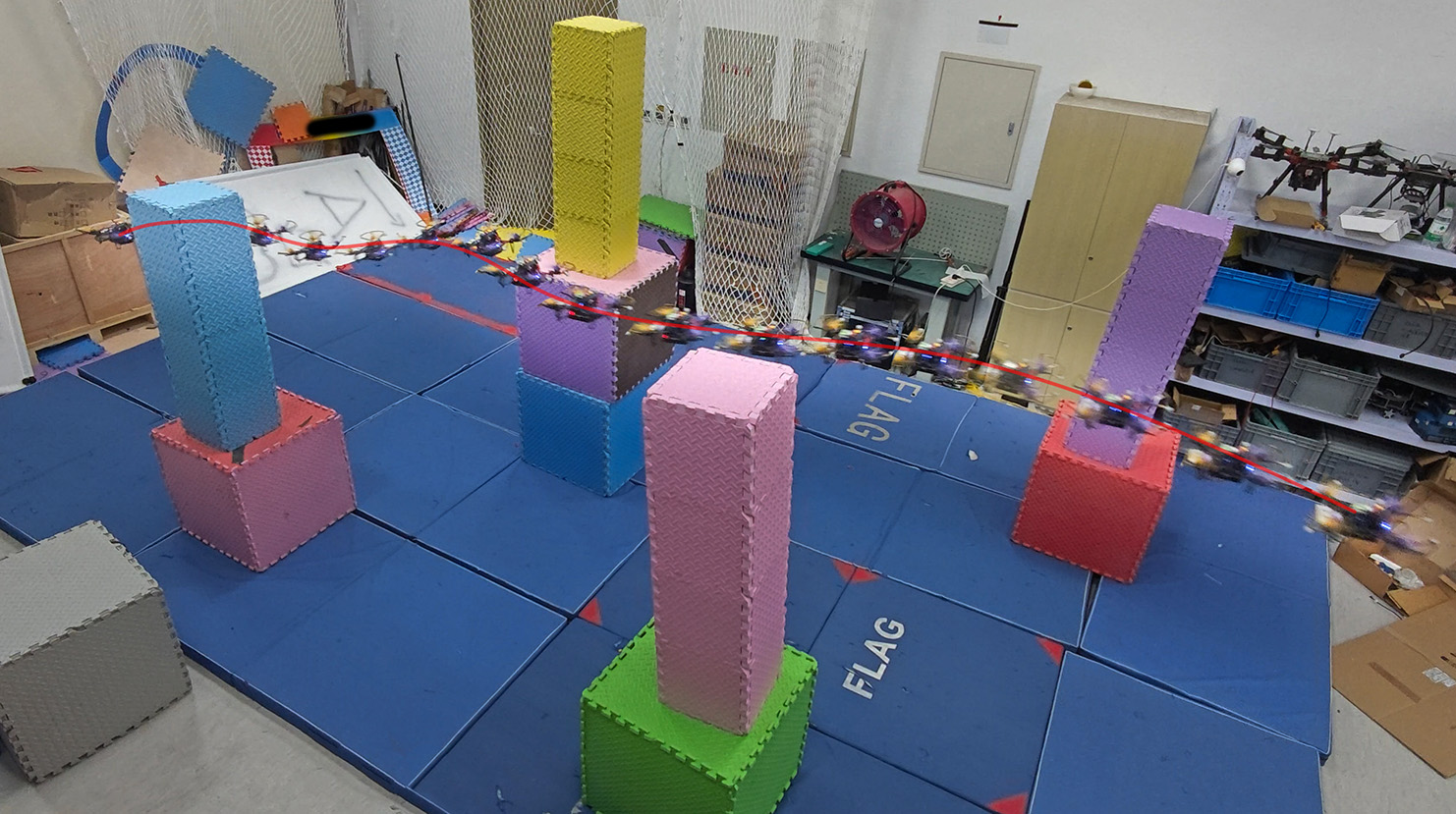}
\caption{Vision-based obstacle avoidance trajectory in the real world.}
\label{realworld}
\vspace{-0.5cm}
\end{figure}

Fig. \ref{deploy_pc} presents the velocity curves of the quadrotor in the position control task in both DiffAero and the real world. With moderate domain randomization, the actual quadrotor’s response closely matches the simulated dynamics. Fig. \ref{deploy_oa} illustrates the flight trajectory of a quadrotor in Gazebo, controlled by a vision-based obstacle avoidance policy. The quadrotor successfully navigates through cluttered corridors, achieving a peak velocity of $4$ m/s. Fig. \ref{realworld} shows the trajectory of a quadrotor in a cluttered real-world environment. The deployment results demonstrate that the trained policies can be seamlessly transferred to other simulators and real-world settings, even in the presence of noisy sensory inputs.

\section{Conclusion and Discussion}
In this paper, we have introduced DiffAero, a fully differentiable quadrotor simulator that harnesses GPU-parallelized physics and custom ray casting to deliver orders-of-magnitude improvements in simulation and rendering throughput compared to existing platforms. Beyond performance, DiffAero provides a modular and extensible research platform that unifies multiple dynamics models, sensor modalities, flight tasks, and learning algorithms within a GPU-native learning interface, thereby enabling systematic investigations of learning-based aerial robotics. To illustrate its capability, we presented a case study using a variant of SHAC (SHA2C) for training vision-based navigation policies with DiffAero. Our extensive benchmarks and real-world experiments demonstrate that DiffAero enables rapid, reliable training and seamless sim-to-real transfer of quadrotor flight policies on standard hardware. 

Looking forward, we expect DiffAero to serve as an enabling tool for addressing fundamental research questions, such as how the choice of dynamics models affects the learning process, the role of non-differentiable reward signals in hybrid learning algorithms, sim-to-real transfer, and multi-agent coordination. We hope that DiffAero will accelerate research in learning-based aerial robotics and foster new advances in agile, autonomous flight control.

\bibliography{reference}
\bibliographystyle{IEEEtran}

\end{document}